\documentclass[10pt,twocolumn,letterpaper]{article}

\usepackage{cvpr}
\usepackage{times}
\usepackage{epsfig}
\usepackage{graphicx}
\usepackage{amsmath}
\usepackage{amssymb}
\usepackage{subcaption}

\usepackage[breaklinks=true,bookmarks=false,colorlinks]{hyperref}

\cvprfinalcopy 

\def\cvprPaperID{****} 

\begin{document}

\title{Taskology: Utilizing Task Relations at Scale}

\author{Yao Lu$^{1\,2}$, S\"oren Pirk$^{1,2}$, Jan Dlabal$^2$, Anthony Brohan$^{1,2}$, Ankita Pasad$^3$\thanks{Work done while at Robotics at Google}, \\Zhao Chen$^4$, Vincent Casser$^4$, Anelia Angelova$^{1,2}$, Ariel Gordon$^{1,2}$\vspace{1mm}\\
$^{1}$Robotics at Google, $^{2}$Google Research, $^{3}$Toyota Technological Institute at Chicago, $^{4}$Waymo LLC\\
\small
 \tt{ \{yaolug, pirk, dlabal, brohan\}@google.com, ankitap@ttic.edu,
 }\\
 \small
 \tt{\{zhaoch, casser\}@waymo.com, \{anelia, gariel\}@google.com}
 
}

\maketitle

\begin{abstract}
Many computer vision tasks address the problem of scene understanding and are naturally interrelated e.g. object classification, detection, scene segmentation, depth estimation, etc. 
We show that we can leverage the inherent relationships among collections of tasks, as they are trained jointly, supervising each other through their known relationships via \emph{consistency losses}. 
Furthermore, explicitly utilizing the relationships between tasks allows improving their performance while dramatically reducing the need for labeled data, and allows training with additional unsupervised or simulated data. We demonstrate a distributed joint training algorithm with task-level parallelism, which affords a high degree of asynchronicity and robustness. 
This allows learning across multiple tasks, or with large amounts of input data, at scale.
We demonstrate our framework on subsets of the following collection of tasks: depth and normal prediction, semantic segmentation, 3D motion and ego-motion estimation, and object tracking and 3D detection in point clouds. We observe improved performance across these tasks, especially in the low-label regime.
\end{abstract}

\vspace{-4mm}
\section{Introduction}

Many tasks in computer vision, such as depth and surface normal estimation, flow prediction, pose estimation, semantic segmentation, or classification, are inherently related 
as they describe the surrounding scene along with its dynamics. While solving for each of these tasks may require specialized methods, most tasks are  connected by the underlying physics observed in the real world. A considerable amount of research aims to reveal the relationships between tasks~\cite{zhou2017unsupervised,casser2019struct2depth,DBLP:journals/corr/abs-1708-07860,Zamir2016Generic3R,DBLP:journals/corr/DharmasiriSD17,TASKONOMY2018,zamir2020robust}, but only a few methods exploit these fundamental relationships. Some approaches rely on the unparalleled performance of deep networks to learn explicit mappings between tasks~\cite{zamir2020robust,TASKONOMY2018}. 
However, while training tasks pairs leverages their relationships, it may lead to inconsistencies across multiple tasks, e.g.~\cite{zamir2020robust}, and points to the alternative of training tasks jointly. 

Multi-task learning targets the problem of training multiple tasks jointly. Common to many approaches is a shared feature-extractor component with multiple ``heads" that perform separate tasks
\cite{DBLP:journals/corr/abs-1708-07860,Zamir2016Generic3R,DBLP:journals/corr/DharmasiriSD17}.
Training multiple tasks together increases the coherency between them and -- in some setups -- also enables their self-supervision~\cite{zhou2017unsupervised,casser2019struct2depth}. However, the joint training also has a few disadvantages. For one, a single model for multiple tasks is difficult to design, maintain and improve, as any changes in the training data, losses, or hyperparameters associated with one of the tasks, also affects all others. 
Secondly, different modalities come with different architectures, which are difficult to merge into a single model. For example, point clouds require sparse processing \cite{qi2017pointnet}, while tensored images use CNNs.
Thirdly, it can become intractable to process a single model -- built to perform multiple tasks -- on a single compute node. 

\begin{figure}[t]
\centering
\includegraphics[width=\linewidth]{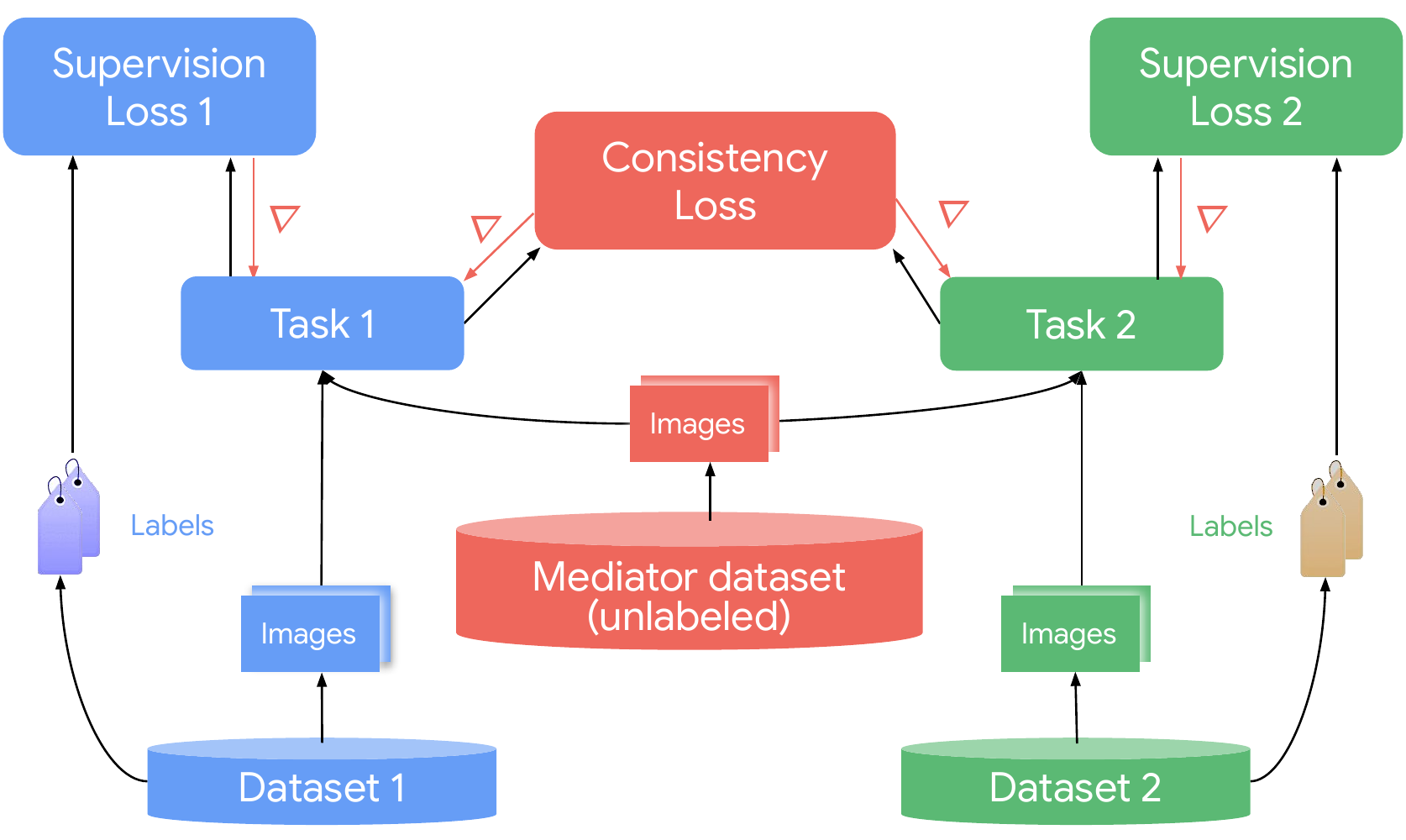}
\caption{Illustration of our framework for the collective training of multiple tasks with a \emph{consistency loss} (two tasks are shown). Each task is performed by a separate network, and trained on a its own dataset and a shared unlabeled \emph{mediator} dataset. The consistency loss is imposed for samples from the mediator dataset.}
\vspace{-3mm}
\label{fig:setup}
\end{figure}

In this paper we introduce a novel approach for distributed collective training that explicitly leverages the inherent connections between multiple tasks (Fig.~\ref{fig:setup}). \emph{Consistency losses} are designed for 
related tasks, intended to enforce their logical or geometric structure. For example, given the two tasks of predicting surface normals and depth from RGB images, the consistency loss between them is based on the analytical relation between them -- normals can be computed from the derivatives of a depth map. We show here that explicitly enforcing consistency between tasks can improve their individual performance, while their collective training also establishes the correspondence among the tasks, which -- in turn -- leads to a more sound visual understanding of the whole scene. 
We term the framework `Taskology', as it connects tasks by their physical and logical constraints.

Using consistency losses to collectively train tasks enables a modular design for training neural networks, which offers three major advantages: We train structurally different tasks with entirely separate networks that are better suited for each individual task. 
This is also advantageous from a design, development, and maintainability point of view; each component can be replaced or improved separately from all others. Secondly, we benefit from \emph{unsupervised or partially labeled data}. For example, many datasets are labeled for either segmentation or scene depth; with consistency losses, we can use partially labeled datasets for training both tasks, where the consistency losses are active for the unlabeled portion of the data (Fig.~\ref{fig:setup}). 
Finally, we train multiple complex models jointly and asynchronously in a \emph{distributed manner}, on different compute nodes. 
Each network is processed on a separate machine, while their training is tied together through consistency losses. The communication between collectively trained networks -- through their predictions -- is asynchronous. Our experiments show that networks for different tasks can be trained with stale predictions from their peers; we do not observe a decrease in performance for up to 20 minute ($\sim2000$ steps) old predictions. Unlike existing methods for distributed training~\cite{mayerjacobsen2020} that mostly rely on data- or model parallelism to split training across multiple compute nodes, our framework separates training at task level; each model is trained independently and asynchronously from all other models, while all models are coherently trained together. Distributed training allows scalability in multi-tasks learning, both in the number of tasks and datasets sizes. 

To summarize, the contributions are:
(1) we present a framework that enables a modular design for training neural networks by separating tasks into modules that can be combined and then trained collectively;
(2) we propose consistency losses for coherently training multiple tasks jointly, which allows improving their overall performance;
(3) we demonstrate distributed training of multiple tasks, which allows for scalability; 
(4) we show that collectively trained tasks supervise themselves, which reduces the need for labeled data, and can leverage unsupervised or simulated data.

\section{Related Work}

Exploiting the structure of -- and between -- tasks has a long tradition in computer science~\cite{turing1950computing,10.5555/94233.94261} and computer vision~\cite{MALIK20164}. 
It has been recognized that knowing about the structure of a task and how it is related to other tasks can be used as a powerful learning scheme~\cite{byravan2017se3,zamir2020robust,TASKONOMY2018,byravan2017se3,zhou2017unsupervised,casser2019struct2depth,godard2017monodepth}. In our work we are interested in making use of these relations more explicitly through the joint co-training of multiple tasks based on consistency losses. Therefore, our method is connected to related work on self- and unsupervised learning, multi-task learning, domain adaptation, and distributed training. While this spans a breadth of related work that we cannot comprehensively discuss, we aim to provide an overview of approaches closest to ours and with a focus on computer vision. 

Methods based on self-supervision aim to autonomously generate labels for training data based on exploiting the inherent structure of related tasks. 
As a prominent example, Doersch et al.~\cite{10.1109/ICCV.2015.167} use unlabeled image collections to learn a representation for recognizing objects. Similarly, many other approaches use proxy or surrogate tasks to learn rich representations for visual data~\cite{10.1007/978-3-319-46466-4_5,pathakCVPR16context,10.1007/978-3-319-46487-9_40,Zamir2016Generic3R,DBLP:journals/corr/abs-1708-06734}. Self-supervised multi-view learning~\cite{zhou2017unsupervised} is closely related to our approach as it aims to train tasks by establishing geometric consistency between them. By designing the consistency losses we directly make use of the known relations between tasks and thereby shape the tasks space, which is similar to common self-supervised training schemes.

The goal of multi-task learning is training models for tasks so as to obtain multiple outputs for a given input, while jointly improving the performance of each individual tasks~\cite{10.5555/3091529.3091535,DBLP:journals/corr/ZhangY17aa,DBLP:journals/corr/Ruder17a}. Many approaches exist that extract features through a shared backbone and then train multiple heads for different objectives~\cite{DBLP:journals/corr/abs-1708-07860,Zamir2016Generic3R}. These approaches are often restricted to tuning the loss function to balance contributions between different tasks~\cite{chen2018gradnorm,kendall2018multi,sener2018multi}. 

While our framework is not limited to any specific task domain or combination of tasks, in this work we focus on the collective training of computer vision models. To this end, we are interested in using established methods for learning depth~\cite{godard2017monodepth,NIPS2014_5539,8575528,zhou2017unsupervised} also together with egomotion~\cite{zhou2017unsupervised,casser2019struct2depth},  surface normals~\cite{DBLP:journals/corr/abs-1903-00112,DBLP:journals/corr/DharmasiriSD17,DBLP:journals/corr/abs-1906-06792,6751533,Qi_2018_CVPR,NIPS2016_6502}, segmentation~\cite{DBLP:journals/corr/Shalev-ShwartzS16,NextSegmPredICCV17,8237857}, optical flow~\cite{Dosovitskiy2015FlowNetLO,hui18liteflownet,10.5555/3298239.3298457,8237343}, or point cloud tracking~\cite{guo2019deep,Ahmed2018DeepLA}. While we do not aim to change the model architectures for any of these tasks, our goal is to improve their performance by designing consistency losses for subsets of these tasks and by jointly training them. 

Finally, the distributed version of our framework is closely related to the concept of distillation~\cite{hinton2015distilling} and online distillation~\cite{anil2018large}, where one network is trained toward an objective with the goal to guide the training of another network. Federated learning \cite{mcmahan17a} is another technique where learning is distributed among multiple instances of the same model. However, unlike them, we explicitly leverage the structure between tasks and by training multiple complex vision tasks simultaneously.

\section{Method}
Our main goal is to enable the distributed and collective training of individual network architectures for computer vision tasks that are inherently related. The main idea is to connect tasks via shared consistency losses, which represent functional or logical relationship. 
In particular, we aim to exploit consistency constraints relating the tasks of predicting depth, surface normals, egomotion, semantic segmentation, object motion, and object tracking with 2D and 3D sensor data. 
For some tasks it is possible to directly formulate their relationship as an analytical differentiable expression -- e.~g.~normals can be computed from the derivatives of depth values \cite{depth2normals}, other tasks are related in more intricate ways, e.~g.~depth and egomotion, or segmentation and optical flow~\cite{Gordon_2019_ICCV,casser2019struct2depth,8237343}.
To train models collectively, we use existing model architectures for specific tasks (e.~g.~such as for predicting depth or segmentation) and define consistency losses between them. In this section we describe our framework, the collective training of two or multiple tasks (Sec.~\ref{sec:cot_setup}), the motivation for and examples of consistency losses (Sec.~\ref{sec:consistency_constraints}), as well as the distributed training of multiple tasks (Sec~\ref{sec:dist_setup}).

\subsection{Collective Training of Tasks}
\label{sec:cot_setup}

Given a set of tasks $\mathcal{T} = \{t_1, \ldots, t_n\}$ 
we define losses for supervising each task individually, which we denote as $\mathcal{L}^{sup}_i$, as well as consistency losses for the collective training of sets of tasks, defined as $\mathcal{L}^{con}$. In the following, tasks are referred to by their index $i$. We then define the overall loss as follows:
\begin{eqnarray} 
\mathcal{L}\!\! &=& \!\!\sum_{i=1}^{n} \mathcal{L}^{sup}_i \bigg(\hat{y}_i(w_i, x), y_i(x)\bigg) \cr \!\!&+&\!\! \mathcal{L}^{con} \bigg(\hat{y}_{1}(w_1, x), \hat{y}_{2}(w_2, x), \dots, \hat{y}_{n}(w_n, x)\bigg),
\end{eqnarray} 
where $\hat{y}_i(\cdot)$ denotes the generated prediction based on the weights $w_i$ of a task $i$, $y_i(\cdot)$ denotes the groudtruth label of a task $i$ and $x$ is a data sample.

We assume that each task is performed by a separate deep network accompanied by a labeled dataset, which we refer to as its \emph{dedicated dataset}. Furthermore, we use the standard supervision loss for each model as if we wanted to train it in isolation. For collective training we then use a separate (unlabeled) dataset, which we refer to as \emph{mediator dataset}, to enforce consistency between the tasks. During training, both tasks receive samples from the mediator dataset and the results of this forward pass are used to compute the consistency loss. The training loop of each task alternates by drawing samples from the dedicated and the mediator dataset. The setup for the collective training of two tasks is illustrated in Fig.~\ref{fig:setup}.
 
The setup described above can have a few special cases: for one, a dedicated dataset for either task, or all, can be empty. In this case the setup reduces to unsupervised training; the unsupervised learning of depth and egomotion 
~\cite{zhou2017unsupervised} exemplifies this case. Datasets can overlap, i.~e.~a dataset can have labels for multiple tasks (e.~g.~for semantic segmentation and depth), or datasets can be only partially labeled, i.~e.~one dataset has labels for depth only and another one has labels for segmentation only. In the latter case the consistency loss will be applied to both datasets, whereas supervised losses are applied to individual datasets only.
The setup naturally generalizes to the collective training of $N$ tasks, where the consistency loss is generally a function of the predictions of all participating tasks. 

\subsection{Task Consistency Constraints}
\label{sec:consistency_constraints}
We rely on the established knowledge \cite{depth2normals,Schonberger_2016_CVPR} in computer vision of tasks and their relationships to identify consistency constraints. Consistency constraints ensure the coherency between different tasks and are derived from laws of geometry and physics. Our goal is to leverage the consistency constraints to define \emph{consistency losses} for task combinations. Any constraint that can be written as differentiable analytic expression can be used within our framework. While this work focuses on already existing relations between tasks via e.~g.~analytical loss relationships, future work can focus on potentially learning these losses. Sections~\ref{sec:seg_depth},~\ref{sec:3d_pc},~\ref{sec:normals} below describe specific task relations considered in this work.

\vspace{-2mm}
\subsubsection{Scene depth, segmentation and ego-motion}
\label{sec:seg_depth}

We first exploit consistency constraints between predicting depth, ego-motion and semantic segmentation. We establish consistency between these three tasks by considering the relations in image pixels and scene geometry between two consecutive frames during training. 
More specifically, we can `deconstruct' a scene at a time-step $t$, estimating its depth and potentially moving objects; at time $t+1$ we can `re-construct' the scene as a function of scene geometry (depth) and moving objects observed at the previous time step and considering the potential ego-motion and objects' motion. Within this setup we impose both geometric and semantics consistencies to reflect the relations between these tasks. More specifically we consider training several tasks: 

\textbf{Motion Prediction Networks:} Given two consecutive RGB frames, $I_1(i, j)$ and $I_2(i, j)$, predict the ego-motion between these frames, i.~e.~the transformation of the camera between frame 2 and frame 1. This can be subdivided into a translation vector $T_{1\rightarrow 2}$ and rotation matrix $R_{1\rightarrow 2}$.
To model independently moving objects, for every pixel $(i, j)$, another task can predict 
the movement of the point visible at the pixel of frame 1, relative to the scene, which occurred between frame 1 and frame 2, denoted as $\delta t_{1\rightarrow 2}(i, j)$. 

\textbf{Depth Prediction Network:} The depth prediction network predicts a depth map, $z(i, j)$, for an image $I(i, j)$.

\textbf{Semantic Segmentation Network:}
From an RGB frame, the semantic segmentation network predicts a logit map $l_c(i, j)$ for each class $c$. For each pixel $(i, j)$, the class is given by $c(i, j) = {\rm argmax}_c\,\, l_c(i, j)$.

The tasks are interrelated as follows: the 3D translation fields can deviate from their background value (due to camera motion) only at pixels that belong to possibly-moving objects (e.g.~vehicles,  pedestrians). Therefore, semantic segmentation informs 3D motion prediction.
Conversely, given a depth map and a 3D motion field,  optical flow fields can be derived and then used to assert the consistency of segmentation masks in pairs of adjacent frames. The flow fields can then be used to inform training of a semantic segmentation module.

Let us define $m(i, j)$ to be the \emph{movable mask}: 
\begin{equation}
    m(i, j) = \left\{\begin{array}{ll}
1 & \quad c(i, j) \in {\cal M} \\
0 & \quad \rm{otherwise} \\
\end{array}\right.
\end{equation}
$\cal M$ is the collection of all classes that represent movable objects, e.g.~persons, cars. For each pixel $(i, j)$, $m(i, j)$ equals 1 if the pixel belongs to one of the movable object classes, and 0 otherwise.

We can now compute the {\it warping} of the first frame onto the second as a result of the scene motion and object motion. 
More specifically, given two adjacent video frames, 1 and 2, a depth map of frame 1, $z_1(i, j)$, the camera matrix $K$, and a pixel position in homogeneous coordinates $p_1(i, j)= (j, i, 1)^T$,
one can write the shift in $p$ resulting from the rotation and translation and object motion $\delta t$ that occurred between the two frames and obtain new values $z_1'(i, j)$ and $p_1'(i, j)$, which are a function of $z$, $p$, $\delta t$ and $m$ and ego-motion network predictions for $R$ and $T$ (see the supp. material) $z_1'(i, j)p_1'(i, j) = KR_{1\rightarrow 2}K^{-1}z_1(i, j)p_1(i, j) +  K (m_1(i, j)\delta t_{1\rightarrow 2}(i, j) + T_{1\rightarrow 2})$. 

Here $p'_1$ and $z'_1$ are respectively the new homogeneous coordinates of the pixel and the new depth, projected onto frame 2. From them we can obtain new estimated values for the image $I_1'(i, j)$, via back projection to the image space. The movable mask $m_1(i, j)$ determines the motion of objects relative to the scene to occur only at pixels that belong to movable objects, i.~e.~$\delta t$ is applied for these objects only.

\paragraph{Consistency Losses:}

The first type of the consistency constraint is \textbf{photometric consistency} across adjacent frames, imposing that RGB values will be preserved after warping. To formulate this constraint, we sample $I_2(i, j)$ at the pixel positions $p'_1(i, j)$, and using bilinear interpolation, we obtain $I_1'(i, j)$, frame 2's RGB image warped onto frame 1. The photometric loss can then be written generally as: 
\begin{equation}
    \mathcal{L}_{\textrm{ph}}\!=\!\sum_{i, j} \!{\cal L}_{\rm p} \!\left(I_1'(i, j), I_1(i, j)\right) + \sum_{i, j}\! {\cal L}_{\rm p}\! \left(I_2'(i, j), I_2(i, j)\right),
\end{equation}
where $I_2'$ is defined analogously to $I_1'$, just with 1 and 2 swapped everywhere. ${\cal L}_{\rm p}$ stands for a pixelwise photometric loss, such as an L1 penalty on the difference in RGB space and structural similarity (SSIM), each weighed by a coefficient. 
In our experiments we used the same photometric loss as described in \cite{Gordon_2019_ICCV}. The depth prediction and motion prediction networks were taken from therefrom as well. The segmentation network was taken from  \cite{Ghiasi_2019_CVPR}.

The second type of the consistency constraint is \textbf{segmentation logits consistency} across adjacent frames. To formulate this constraint, we sample $l_{c2}(i, j)$ at the pixel positions $p'_1(i, j)$, and using bilinear interpolation, we obtain $l_{c1}'(i, j)$. The segmentation consistency loss can then be written generally as: 
\begin{eqnarray}
    \mathcal{L}_{\textrm{seg}}&=&\sum_{i, j, c} {\cal L}_2 \left(l_{c1}'(i, j), l_{c1}(i, j)\right) +\cr &+& \sum_{i, j, c} {\cal L}_2 \left(l_{c2}'(i, j), l_{c2}(i, j)\right),
    \label{eq:maskcon}
\end{eqnarray}
where $l'_{c2}$ is defined analogously to $l_{c1}'$, just with 1 and 2 swapped everywhere. ${\cal L}_2$ stands for a L2 loss squared.

Overall, the final consistency loss becomes 
\begin{equation} 
\mathcal{L}_{2D\_sem}^{Con}=\mathcal{L}_{\textrm{ph}}+\mathcal{L}_{\textrm{seg}}.
\end{equation}


\begin{figure}
\centering
\includegraphics[width=0.99\linewidth]{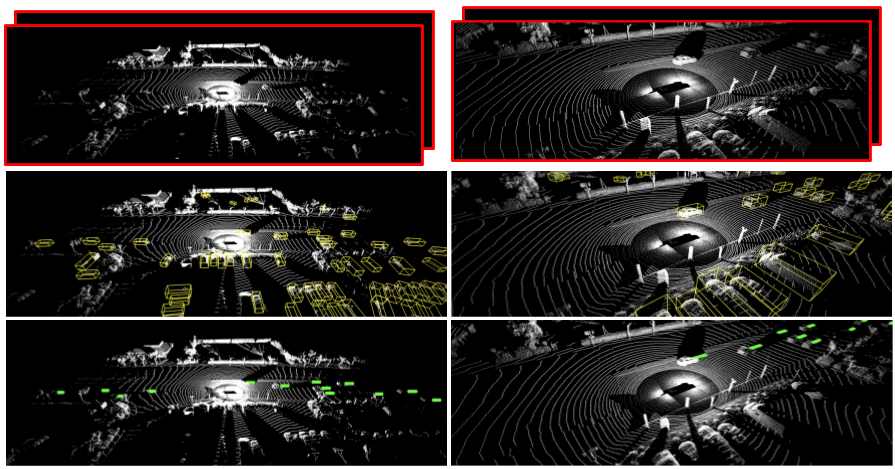}
\caption{3D Object Detection setup: from a sequence of point clouds (top), we predict bounding boxes per individual point cloud (middle) and the optical flow, denoted as green arrows (bottom).}
\label{fig:3dpc}
\end{figure}

\subsubsection{3D Object Detection in Point Clouds in Time} 
\label{sec:3d_pc} 

In this section, we focus on object detection from 3D point clouds, a difficult task which also is crucial in many practical applications such as for autonomous vehicles. We show that, when we simultaneously train an object flow network, which predicts motion of objects through time, we can apply a motion-consistency loss to significantly boost the single-frame 3D object detector performance, especially in the low-label regime. 
Similarly to establishing consistencies between dynamic scenes in 2D space as in Section~\ref{sec:seg_depth}, one can establish consistencies in 3D. More specifically we can track moving objects in 3D across frames. Given a module that detects movable objects in a scene, in two consecutive frames, and another module that predicts rigid motion, we can assert that the motion-prediction module correctly estimates the motion of each object. We here directly enforced this in 3D space using 3D bounding box detections in Point Clouds and optical flow.

We address 3D Object Detection in Point Clouds using two models: one predicting 3D bounding boxes on a point cloud, and the other predicting 2D box flow on point cloud sequences (Figure~\ref{fig:3dpc}). 
We use a PointPillar-based \cite{lang2019pointpillars} network as our 3D detector, which allows us to work in a pseudo-2D top-down space for all our experiments. At the shared feature layer preceding the detector prediction head we attach a flow prediction head that, given $n_f$ frames, outputs $3(n_f-1)$ channels corresponding to flow. 
Our detection model operates on the grid-voxelized input point cloud of shape $(n_x, n_y)$, where we choose $n_x=n_y=468$ (the $z$ dimension has been marginalized out as we are using a PointPillar~\cite{lang2019pointpillars} model). The grid size in $x-y$ corresponds to each grid point being of extent (0.32m, 0.32m) in real space. For each grid point, we predict:

\begin{enumerate}
\vspace{-1mm}
\item $(7n_a, n_f)$ residual values pinned to $n_a$ fixed anchors. The 7 values correspond to displacements $dx, dy, dz, dw, d\ell, dh, d\theta$ of the final predicted box from the values for the anchor boxes. Ground truth boxes are automatically corresponded to anchors at training time. 
\item $(n_a, n_f)$ class logit values denoting confidence that an object of the specified type exists for that anchor box.
\item $(n_a, 3(n_f-1))$ values corresponding to the box flow $(flow_x, flow_y, flow_{\theta})$ of a hypothetical box in the current frame to any $n_f-1$ frames in the past. 
\end{enumerate}

As mentioned we use the same backbone to predict box flow. Namely, through an equivalent backbone we predict a three-channel map 
$(\textrm{flow}_x, \textrm{flow}_y, \textrm{flow}_{\theta})$
of the same resolution of the detector predictions, corresponding to the flow of the boxes in a current frame to any of the previous frames in the sequence. The flow is only supervised at locations within the grid associated with positive object detections. 

\textbf{Consistency Loss:} This set of detection and flow predictions induces natural consistency constraints. Namely, given a predicted flow that transforms an anchor point centered at $(x,y)$ in the current frame to the closest anchor point 
$(x+\textrm{flow}_x, y+\textrm{flow}_y, \theta+\textrm{flow}_{\theta}) \approx (x',y',\theta')$ in another frame, we consider the following two loss functions (in the following, all primed coordinates are coordinates after flow has been applied to the current frame): 

\begin{enumerate}
    \item $\mathcal{L}_{\textrm{class}}$: The class confidences at two points $(x,y,z) \mapsto (x+\textrm{flow}_x,y+\textrm{flow}_y,\theta+\textrm{flow}_\theta)$ connected by the predicted flow vector should have the same class confidence. $\mathcal{L}_{\textrm{class}} = \left(\textrm{classlogit}(x,y) - \textrm{classlogit}(x',y')\right)^2$
    \item $\mathcal{L}_{\textrm{residual}}$: The predicted flow can be used to calculate consistent residual values for $(x,y,\theta)$ at two points connected by the predicted flow. The residual values for $z,w,\ell$, and $h$ should also remain constant between two detections connected by flow (for short time spans we assume near-constant elevation). $\mathcal{L}_{\textrm{residual}} = \sum_{i\in (x,y,\theta)}(di' - di + (\textrm{flow}_i-(i'-i))^2 + \sum_{j\in (z,\ell,w,h)}(dj' - dj)^2$
\end{enumerate}
Overall, our 3D Point Cloud motion-consistency loss becomes $\mathcal{L}_{PC\_in\_time}^{Con}=\mathcal{L}_{\textrm{class}}+\mathcal{L}_{\textrm{residual}}$. 

The class loss $\mathcal{L}_{\textrm{class}}$ ensures that class logits along predicted object tracks are equal, while the residual loss $\mathcal{L}_{\textrm{residual}}$ do the same for residual values along tracks. The first term in the residual consistency takes into account the predicted flow, which transforms $(x,y,\theta)$ to $(x',y',\theta')$. Because the flow is continuous and not quantized like the voxel grid, we can normalize out the quantization noise exactly by adding a remainder term, $(\textrm{flow}_i-(i'-i))$ for $i\in (x,y,\theta)$. For the residual consistency, we further enforce that the bounding box residuals for $z,\ell, w, h$ do not change along object tracks. This reflects our assumption that the dimensions of the vehicle are preserved and there is no appreciable elevation change over the span of a second.

\subsubsection{Depth and Surface Normals}
\label{sec:normals}

\begin{figure}
\centering
\includegraphics[width=0.9\linewidth]{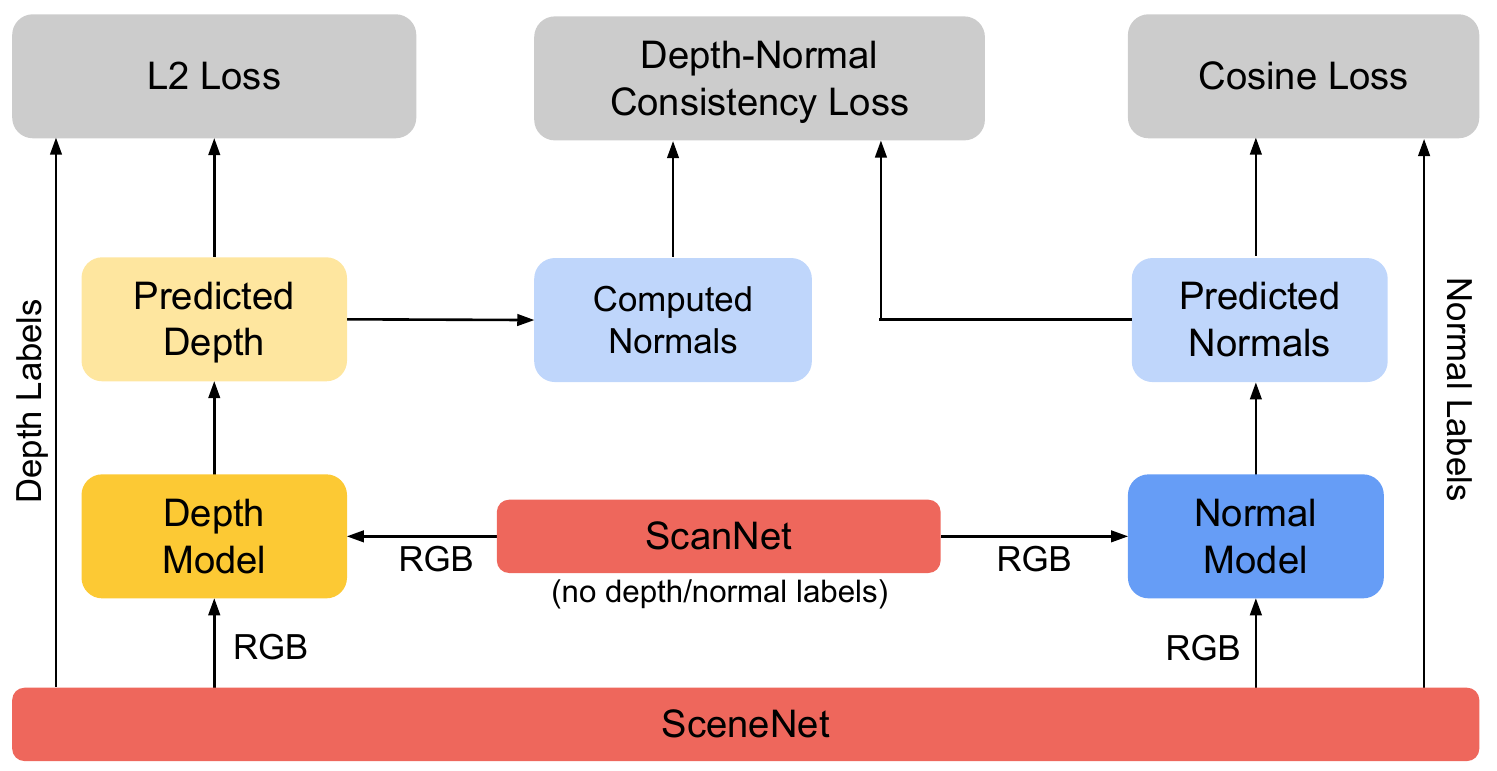}
\caption{Depth and normals joint training for domain adaptation: we use SceneNet (simulated) data to supervise the training of separate models for depth and surface normal prediction and apply a consistency loss to jointly train both models for images from ScanNet, where we do not use any ground truth normals labels.}
\vspace{-2mm}
\label{fig:domain_adapt}
\end{figure}

Given a depth prediction module and a surface-normal prediction module, we can assert that the normals obtained from the spatial derivatives of the depth map are consistent with the predicted normals.  
More specifically, the depth model can produce continuous depth, from which one can analytically compute surface normals estimates per each location $\hat{n_d}$, which are a function of depth. On the other hand, a surface normals model, can be trained independently to produce surface normals predictions $\hat{n_p}$, and a consistency loss between these two predictions of surface normals can be imposed on the shared data source (Figure~\ref{fig:domain_adapt}).
The consistency is then computed as: 
\begin{equation}
    \mathcal{L}_{Normals}^{Con} = \textrm{cosine\_distance}(\hat{n_d}, \hat{n}_p),
\end{equation}
where $\hat{n_d}$ is the computed surface normals from the inferred depth and $\hat{n_p}$ is the normal map predicted from the normal prediction network (see the supp. material for derivation).

Interestingly, that can also be done by combining simulated and real data sources.
Figure~\ref{fig:domain_adapt} visualizes the setup used in our experiments later, where the data source for surface normals (supervised) training is simulated, whereas a real data source can be used by both (it is used in unsupervised manner for the normals model).

\subsection{Distributed Training}
\label{sec:dist_setup}
Collective training of multiple networks eventually requires distributing the computation across multiple compute nodes, to speed up the training, or simply because a large enough collection of models cannot be processed on a single machine. In distributed training it is often the communication between the nodes that sets the limitations \cite{mayerjacobsen2020}. 
Our framework provides the advantage of training tasks independently, with communication via consistency losses only. Furthermore, the modules share, potentially vast amounts of unsupervised, data, which allows for data-parallelism.

To reduce the communication load, distributed training schemes often aim to be asynchronous, which means that model parameters or their gradient updates develop some degree of `staleness', which denotes the interval between updates to each model.
It is easy to observe that stale predictions are less harmful than stale gradient updates~\cite{anil2018large}, since predictions are expected to converge as the training progresses. 
Therefore, modules will query each other's predictions to compute shared losses, but propagate gradients locally, within their own module.  

\begin{figure}[t]
\centering
\includegraphics[width=\linewidth]{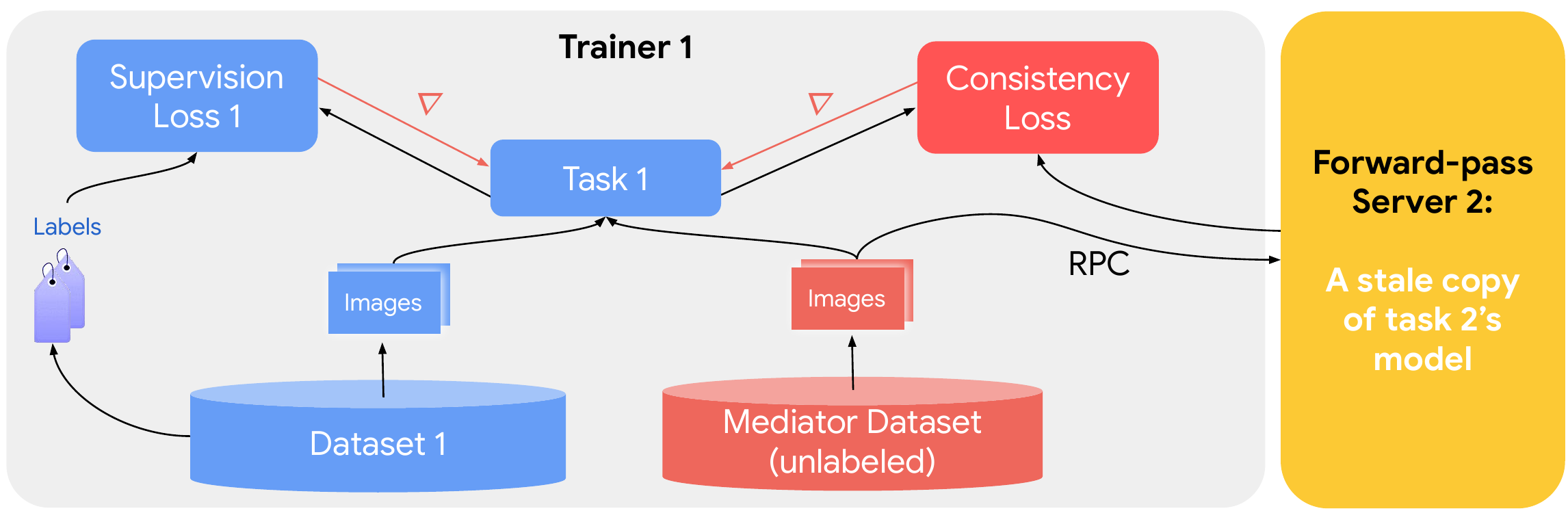}
\caption{Illustration of our distributed setup for collective training. Each task-module is training on its own machine (its ``trainer", only Trainer 1 is shown in the figure). In order to compute the consistency loss, Trainer 1 reaches out to a server that hosts a stale copy of Task 2's model and performs the forward pass (and vice versa). Each trainer pushes gradient updates to its respective model, and every so often, the stale copies on both forward-pass servers are refreshes with fresh copies form the respective trainer.}
\vspace{-5mm}
\label{fig:dist_setup}
\end{figure}

Our distributed implementation is based on this principle and takes advantage of shared losses. 
Each module is training on a separate machine (``trainer"), as illustrated in Fig.~\ref{fig:dist_setup}. The consistency loss depends on the outputs of all co-training tasks, which means that its computation requires evaluating a forward pass through all of them. Each trainer evaluates the forward pass of its own module and, to evaluate the forward passes of the other modules, it queries servers that host stale copies of the peer modules. At each training step, the trainer then pushes gradients to its own module and updates its weights. 

One advantage of our distributed method is that each module can train with its own hyperparameters, including optimizer, regularizers, and learning rate schedules. Typically per-task modules are published together with these hyper-parameters, and our method allows using them as necessary for each respective model. Moreover, more computationally expensive modules can be allocated with more computational resources, to approximately equalize the training times among the modules. Finally, since the modules communicate through predictions, and predictions are typically much more lightweight than network weights, the communication overhead is significantly lower compared to other distributed training techniques.

\section{Experiments}

In the following sections we report results of experiments on using consistency losses, 
for the co-training of multiple models as described in Sections~\ref{sec:seg_depth},~\ref{sec:3d_pc},~\ref{sec:normals}. We observe improved performance across tasks (Section~\ref{sec:3tasks}), successful training with unlabeled data, where our approach is more helpful in lower label regimes (Section~\ref{sec:pointclouds}), and successful domain adaptation (Section~\ref{sec:normals_shift}). 
The experiments in Sections \ref{sec:3tasks}, \ref{sec:pointclouds}, and \ref{sec:staleness} were run distributed, whereas the rest of the experiments were run on a single machine (e.~g.~as in Fig.~\ref{fig:setup}).

\subsection{Scene Depth, Segmentation and Ego-motion} \label{sec:3tasks} 
For this experiment we show results for the distributed collective training of three tasks. The first task is semantic segmentation based on NAS-FPN \cite{Ghiasi_2019_CVPR}. The other two tasks are depth prediction and motion estimation (for both camera and objects), for which we rely on existing models~\cite{Gordon_2019_ICCV}. Segmentation masks were used to regularize the 3D motion fields~\cite{Gordon_2019_ICCV} (Section~\ref{sec:seg_depth}). 
Semantic segmentation was trained on COCO \cite{DBLP:journals/corr/LinMBHPRDZ14} 2017 as its dedicated dataset and Cityscapes~\cite{Cordts2016Cityscapes} was used as the unlabeled mediator dataset -- no Cityscapes labels were used at training. Each of the three models was trained on a separate machine, as outlined in Sec.~\ref{sec:dist_setup}. The segmentation module received a greater allocation of compute resources than the others, since it is significantly more computationally expensive. The hardware configuration is described in the supp. material. The batch size, the optimizer, the learning rate, and other hyperparameters varied across the tasks and based on the respective published values for each model. During training, each of the three models queries its peers via RPC to obtain their predictions, which were up to \emph{one minute} stale. 

\begin{table} [h!]
\small
\centering
\scalebox{0.72}{
  \begin{tabular}{|l|c|c|c|}
  \hline
   &  Depth Error & Segmentation   \\
  Configuration & (Abs. Rel.) & MIOU \\
  \hline
  A. Depth \& motion only & 0.165 &- \\
  B. Segmentation only&-   & 0.455\\
  C. Frozen segmentation model B with depth \& motion  & 0.129 &-  \\
  D. Frozen depth \& motion model C \& segmentation &-  & 0.471 \\
  E. Depth, motion and segmentation training jointly  & \textbf{0.125}& \textbf{0.478}\\
    \hline
  \end{tabular}

}
\vspace{0.5mm}
\caption{\small Results of the distributed collective training of three models: Depth prediction, 3D motion prediction, and semantic segmentation. COCO was the dedicated dataset for semantic segmentation, and Cityscapes served as an unlabeled mediator dataset. 
Both depth prediction and segmentation were evaluated on Cityscapes, with segmentation evaluated only for predictions associated with pedestrians and vehicles (details of the evaluation protocol are given in the supp. material).
}
    \label{tab:3workers}
\end{table}

The effect of collective training on the performance of the participating models is shown in Tab.~\ref{tab:3workers}. Experiments A and B are the baselines, where the depth and motion models were trained jointly, but separately from segmentation. Experiment E shows the improvement in performance when all three tasks train jointly with consistency constraints. Rows C and D are ablations that demonstrate the changes in performance when consistency constraints are turned on progressively. In C the depth and motion models are supervised by the segmentation model from experiment B. C achieves the same depth error as a similar configuration trained on a single machine~\cite{Gordon_2019_ICCV}, where segmentation masks were precomputed. In experiment D, segmentation was concistency supervised by the improved depth and motion model from experiment C, but the latter two models remained frozen. The progression in quality demonstrates the effect of consistency supervision on all tasks.

While consistency contributes to correctness, it does not guarantee the latter. This is reflected in the failure cases of the method. Some illustrative examples are shown in the supp. material.

\subsection{3D Object Detection in Point Clouds in Time}
\label{sec:pointclouds}

We perform all experiments on the vehicle class of the Waymo Open Dataset \cite{sun2019scalability} which provides complete tightly-fitting 3D bounding box annotations along with tracks for each vehicle and use a sequence length $n_f=3$ point clouds ($\Delta=0.5s$). 
Our backbone architecture is based on a PointPillar detector \cite{lang2019pointpillars}. Given a sequence of input point clouds, we quantize the points for each frame into a grid in the $x$-$y$ plane and then use our single-frame detection model to produce a confidence value at each grid point for the presence of an object box as well as residual values $(x,y,z,w,\ell,h,\theta)$ to refine the final box coordinates (Section~\ref{sec:3d_pc}).
For all our reported experiments, $n_f=3$ and $n_a=2$. 
We follow the original PointPillar network settings in choosing all class thresholds. We perform experiments with partial labeling in which only 5\% or 20\% of the box labels are available. Our baseline (100\%) is trained in isolation, whereas all partial label studies are performed in the distributed framework.

\begin{table} [t]
  \begin{center}

  \scalebox{0.75}{
  \begin{tabular}{|l|c|c|c|}
  \hline
  Method & Labels & 3D mAP/mAPH (\%) & BEV mAP/mAPH (\%)  \\
  \hline
  
  No Consistency & 5\% & 17.6/9.6 & 44.3/24.3\\

  Adding $\mathcal{L}^{con}$ & 5\% & \textbf{23.5}/\textbf{12.0} & \textbf{51.1}/\textbf{26.5}   \\

  \hline
  
  No Consistency & 20\% & 30.8/16.4 & 63.0/34.1\\

  Adding $\mathcal{L}^{con}$ & 20\% & \textbf{31.6}/\textbf{19.1} & \textbf{65.7}/\textbf{39.2} \\
  
  \hline
  
  No Consistency & 100\% & 53.0/47.6 & \textbf{75.0}/66.8 \\

  Adding $\mathcal{L}^{con}$ & 100\% & \textbf{54.2}/\textbf{49.6} & \textbf{75.0}/\textbf{68.5} \\

  \hline
 
  \end{tabular}
     
    }
  \vspace{0.5mm}
 
        \caption{3D detection and 2D (BEV) metrics on the Waymo Open Dataset, given various degrees of dataset labels provided for training. We can see consistent improvements when applying our motion-based consistency loss, especially with fewer labels.} 
  \vspace{-6mm}
    \label{tab:track_pred}
    \end{center}

\end{table}
Our results are shown in Table \ref{tab:track_pred}. 
We can see the three sets of experiments, in which we stripped the dataset of its labels to various degrees. Our metrics are based on the standard mean average precision (mAP) metrics for 2D and 3D detection. We also use the mAPH metric introduced in \cite{sun2019scalability}, which takes into account object heading. mAPH is calculated similarly to mAP, but all true positives are scaled by err$_{\theta}/\pi$, with err$_{\theta}$ being the absolute angle error of the prediction in radians. We also report results on both 3D detection and 2D Bird-Eye-View (BEV) detection.

We can see that joint training with consistency losses is very beneficial. 
The consistency loss improves the object detector performance in all three settings, with more significant improvements when labels are scarce. This also holds for both 3D detection and 2D BEV detection. Furthermore, the consistency loss has a beneficial effect on mAPH, i.e. is able to correct errors in heading, as it enforces rotational consistency along each object track.

\subsection{Depth and Surface Normals with Domain Shift}
\label{sec:normals_shift}
Since training in our framework involves multiple datasets, it is interesting to explore what happens when there is a large domain disparity between them. To this end, we select the extreme case of domain disparity between the dedicated datasets and the mediator dataset. As tasks for this experiment we selected depth estimation and surface normal prediction. We use SceneNet~\cite{McCormac:etal:ICCV2017}  (simulated data) as the dedicated dataset, and ScanNet~\cite{dai2017scannet} (real data) as the unlabeled mediator dataset (Figure~\ref{fig:domain_adapt}). We use simulated SceneNet to train depth and normal estimation models and evaluate them on the real ScanNet data as our baseline. The strong domain disparity is evident from the fact that a model trained on simulated data performs poorly on the real dataset (Table~\ref{tab:depth_normal_exp}). 
For the baseline both models are trained separately to predict depth and surface normals, with a mean squared error loss for the depth model and a cosine loss for the surface normal model. The trained models are then used to predict surface normals for samples from ScanNet. 
Accuracy is measured by using the ground truth data of ScanNet for depth and surface normals generated by the method of~\cite{DBLP:journals/corr/abs-1906-06792}.

We then train the models with consistency loss on ScanNet. The consistency loss can then be computed as cosine similarity of the computed surface normals and those predicted by the normal prediction network. The consistency is based on the fact that a normal map can be analytically computed from a depth map ~\cite{5152493} (Section~\ref{sec:normals}). 

Table~\ref{tab:depth_normal_exp} shows the results of individual training of depth and surface normals prediction on SceneNet (simulated) and tested on ScanNet (real), and when training in the same transfer setting but with loss consistency. We observe that training with loss consistency improves the performance on both tasks on this challenging sim-to-real transfer task. 

\begin{table}[h]
\begin{center}
\scalebox{0.74}{
\begin{tabular}{|l|c|c|c|c|}
 \multicolumn{1}{c}{} &  \multicolumn{3}{c}{Normals} & \multicolumn{1}{c}{Depth}\\
 \multicolumn{1}{c}{} &  \multicolumn{3}{c}{Accuracy (in \%)} & \multicolumn{1}{c}{Error (in \%)}\\
 \hline
Method &   $<11.25^{\circ}$ & $<22.50^{\circ}$ & $<30.00^{\circ}$ & Abs. Rel\\
\hline
SceneNet $\rightarrow$ ScanNet & 9.2 & 30.8 & 46.3 & 28.2\\
SceneNet $\rightarrow$ ScanNet &     &      &      &  \\ (with Consistency) & \bf 13.6 & \bf 34.9 & \bf 46.7 & \bf 24.9 \\
\hline
\end{tabular}
}
\end{center}
\caption{\small Surface normal prediction transfer from SceneNet (simulated) to ScanNet (real).}
\vspace{-4mm}
\label{tab:depth_normal_exp}
\end{table}

\subsection{Tolerance to Staleness}
\label{sec:staleness}

As discussed in Sec.~\ref{sec:dist_setup}, in our setup, individual modules communicate with each other through their predictions. This is motivated by the increased resilience to staleness that predictions exhibit compared to weights and gradients~\cite{anil2018large}. To study the amount of staleness our setup can afford, we train depth and egomotion~\cite{casser2019struct2depth} on the KITTI dataset~\cite{Geiger2013IJRR}, each on a separate machine. This experiment is particularly challenging because it is fully unsupervised: each of the modules is only supervised by the predictions produced by its peer. Since both modules are randomly initialized, each model initially receives a \emph{random} and \emph{stale} peer-supervision signal.

\begin{figure}[h]
\centering
\includegraphics[width=0.99\linewidth]{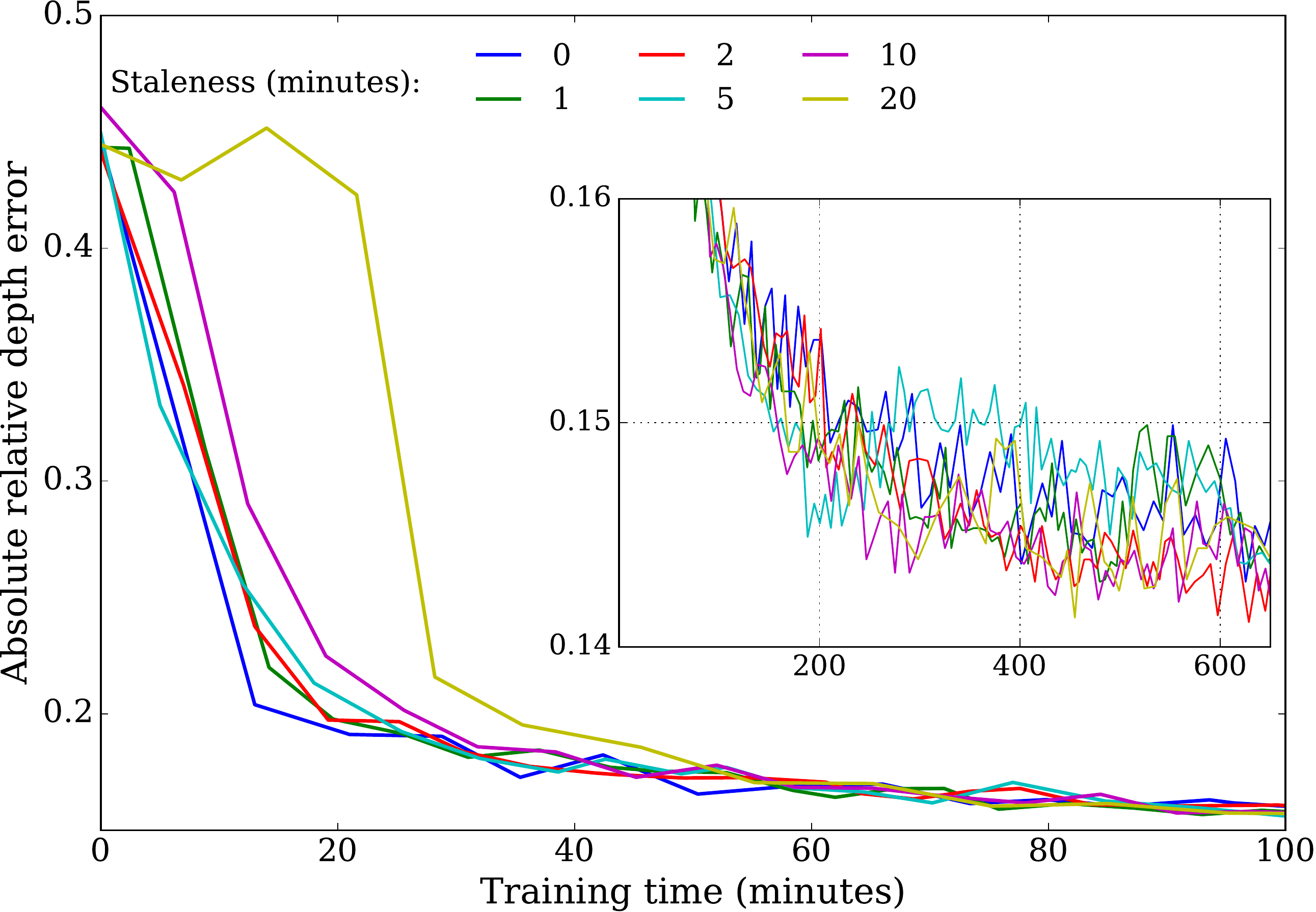}
\caption{Depth prediction error as function of time for different values of staleness for distributed collective training on depth and egomotion on KITTI. Staleness of 20 minutes means that the depth trainer receives egomotion labels from an egomotion model that refreshes every 20 minutes, and vice versa. The depth trainer performs about 100 training steps per minute, so \emph{20 minutes} translates to \emph{2000 steps}. A value of 0 staleness denotes a configuration where all networks were placed on the same machine and trained synchronously. The graphs in the inset show the long-time progression of training. All experiments achieve the same absolute relative depth prediction error of about 0.143 (which on is par with the state-of-the-art~\cite{casser2019struct2depth} for models that disregard object motion), at about the same time, irrespectively of the staleness.}
\label{fig:async}
\end{figure}

 In Fig.~\ref{fig:async} we show the results of this experiment, which is the depth prediction error as function of time for various values of staleness. While greater staleness values initially hinder the training, all experiments converge to approximately the same result, and approximately at the same time. Staleness of up to \emph{20 minutes} -- or \emph{2000 training steps} -- is shown to have no adverse effect on the convergence time or the test metric.

The negative effects of staleness on convergence time and on result metrics have been studied for various distributed training methods \cite{dai2018understanding,chen2016revisiting,chen2018efficient,dutta2018slow}. While the tolerance to staleness varies widely, due to the diversity of methods, most studies only report the sensitivity of methods to staleness of up to a few tens of steps. Unlike these findings, our distributed training setup is much more robust and thereby enables training with staleness of up to thousands of steps.

\section{Conclusions}
We have introduced a novel framework, `Taskology', for the collective training of multiple models of different computer vision tasks. Our main contribution is that our framework enables a modular design for training neural networks by separating tasks into modules that can be combined and trained collectively. Furthermore, we employ consistency losses so as to exploit the structure between tasks. By jointly training multiple tasks, we have shown that consistency losses help to improve the performance, and can take advantage of unlabeled and simulated data. 
Our approach achieves better results from joint training, especially when a large portion of the dataset is not labeled.
We also demonstrated a distributed version of the framework, which trains models on separate machines and is robust to staleness. 

{\small
\bibliographystyle{ieee_fullname}
\bibliography{egbib}
}

\newpage
\section*{Supplementary Material}

\section{Scene depth, segmentation and ego-motion} \label{sec:dms}
\subsection{Modules and Interfaces}
The interfaces of the three modules, depth, motion and semantic segmentation, are defined below:

\paragraph{Motion Prediction Network:} Given two consecutive RGB frames, $I_1(i, j)$ and $I_2(i, j)$, of width $w$ ($0 \leq j < w$) and height $h$ ($0 \leq i < h$), the motion prediction network predicts the following quantities: 
\begin{itemize} 
\item $\delta t_{1\rightarrow 2}(i, j)$: For every pixel $(i, j)$, $\delta t_{1\rightarrow 2}(i, j)$ estimates the movement of the point visible at the pixel $(i, j)$ of frame 1, relative to the scene, which occurred between frame 1 and frame 2.
\item $T_{1\rightarrow 2}$: The translation vector of the camera between frame 2 and frame 1.
\item $R_{1\rightarrow 2}$: The rotation matrix of the camera between frame 2 and frame 1.
\end{itemize}
Similarly, the network predicts $\delta t_{2\rightarrow 1}(i, j)$, $T_{2\rightarrow 1}$, and  $R_{2\rightarrow 1}$, which are defined as above, with (1) and (2) swapped. 

\paragraph{Depth Prediction Network:}
Given an RGB frame, $I(i, j)$, the depth prediction network predicts a depth map, $z(i, j)$, for every pixel $(i, j)$.

\paragraph{Semantic Segmentation Network:}
Given an RGB frame, $I(i, j)$, the semantic segmentation network predicts a logit map $l_c(i, j)$ for each class $c$. For each pixel $(i, j)$, the class is given by $c(i, j) = {\rm argmax}_c\,\, l_c(i, j)$.

\subsection{Next frame warping}
To construct the consistency losses for these tasks, as shown in the main paper, we need to derive the locations of each pixel from the first frame onto the next frame, which is also referred to as image warping from frame 1 to frame 2. 
We start with defining $m(i, j)$ to be the \emph{movable mask}: 
\begin{equation}
    m(i, j) = \left\{\begin{array}{ll}
1 & \quad c(i, j) \in {\cal M} \\
0 & \quad \rm{otherwise} \\
\end{array}\right.
\end{equation}
$\cal M$ is the collection of all classes that represent movable objects. These are detailed below. For each pixel $(i, j)$, $m(i, j)$ equals 1 if the pixel belongs to one of the movable object classes, and 0 otherwise.

Given two adjacent video frames, 1 and 2, a depth map of frame 1 $z_1(i, j)$, the camera matrix $K$, and a pixel position in homogeneous coordinates
\begin{eqnarray}\label{eqp}
p(i, j) = \begin{pmatrix}  
     j  \\ i \\  1 
     \end{pmatrix},
\end{eqnarray}
one can write the shift in $p$ resulting from the rotation and a translation that occured between the two frames as:
\begin{eqnarray}\label{warp_expand}
z_1'(i, j)p_1'(i, j) &=& KR_{1\rightarrow 2}K^{-1}z_1(i, j)p_1(i, j) \cr &+&  K (m_1(i, j)\delta t_{1\rightarrow 2}(i, j) + T_{1\rightarrow 2}),
\end{eqnarray}
where $p'_1$ and $z'_1$ are respectively the new homogeneous coordinates of the pixel and the new depth, projected onto frame 2, and $K$ is the camera matrix. 
The above equation consists of the scene depth, as obtained by rigid motion of the scene and the additional changes obtained from the motions of the individually movable objects. Note that the motion mask is only applied to regions of potentially movable objects  $m_1(i, j)$, determined by the semantic segmentation model.
The movable mask $m_1(i, j)$ (of frame 1) restricts motion of objects relative to the scene to occur only at pixels that belong to movable objects.

\subsection{Evaluation Protocol} \label{sec:classes}
In our experiment COCO served as the dedicated dataset for segmentation, and Cityscapes served as the unlabeled mediator dataset. Since the two datasets have different sets of labels, we had to create a mapping between the two. The mapping is shown in Table \ref{tab:labels}. 

\begin{table}[]
\begin{center}
\begin{tabular}{|l|c|c|c|}
\hline
 & \multicolumn{3}{c|}{Label ID}  \\ \hline
Class\,\, & \,Ours\, & COCO 2017 & Cityscapes \\ \hline
person/rider  \,  & 1            & 1                           & 24/25              \\ 
bicycle         & 2            & 2                           & 33                 \\ 
car             & 3            & 3                           & 26                 \\ 
motorcycle      & 4            & 4                           & 32                 \\ 
traffic lights  & 5            & 10                          & 19                 \\ 
bus             & 6            & 6                           & 28                 \\ 
truck           & 7            & 8                           & 27                 \\ \hline
others          & 8            & \,other labels \,            & \, other labels\,    \\ \hline
\end{tabular}
\vspace{1mm}
\caption{Mapping between Cityscapes label IDs, COCO labels IDs, and the label IDs we defined for this experiment.}
\vspace{2mm}

\label{tab:labels}
\end{center} 
\end{table}
Only labels that represent movable objects are of interest for our experiment. We therefore restricted our label set to 7 classes, that are in the intersection of Cityscapes and COCO and represent movable objects. All other labels were mapped to label ID 8. When evaluating the segmentation on Cityscapes, we mapped the Cityscapes groundtruth labels and the COCO-trained model predictions to these 8 labels.

\subsection{Hardware configuration}

The three models in this experiment had different computational costs. Table \ref{tab:step_time} shows the duration of a training step for each of the three models on 8 NVIDIA p100 GPU, for a batch of 32. Placing the segmentation model on a TPU node reduced its training step time to be closer to the other modules. This way the convergence was not gated on the Segmentation model. Being able to train each module on a different hardware configuration is one of the strengths of our method.

\begin{table}[htb]
\begin{center}
\begin{tabular}{|l|c|c|}
\hline
Model & Hardware & Step time \\
 \hline
Depth & GPU & 0.81s \\
Motion & GPU & 0.83s \\
Segmentation & GPU & 2.18s \\
Segmentation & TPU & 1.42s \\
\hline
\end{tabular}
\vspace{2mm}
\caption{Time per training step in milliseconds for each of the three modules in Sec.~3.2.1 in the main paper, on various hardware platforms. The batch size is 32 in all cases. GPU denotes 8 NVIDIA p100 GPU, and TPU denotes a Google Cloud TPU v2-8 unit.}
\label{tab:step_time}
\end{center} 
\end{table}

\subsection{Failure cases}
\begin{figure} [t]
\centering
\begin{subfigure}{0.5\textwidth}
\includegraphics[width=\linewidth]{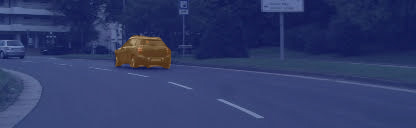}
\caption{Failure example 1, frame 1}
\end{subfigure}
\hspace*{\fill}
\begin{subfigure}{0.5\textwidth}
\includegraphics[width=\linewidth]{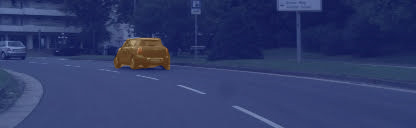}
\caption{Failure example 1, frame 2}
\end{subfigure}
\caption{Failure example 1: Segmentation network fails to segment out a white car on the left edge on two consecutive frames.}
\label{fig:1_overlay}
\end{figure}

\begin{figure} [t]
\centering
\begin{subfigure}{0.5\textwidth}
\includegraphics[width=\linewidth]{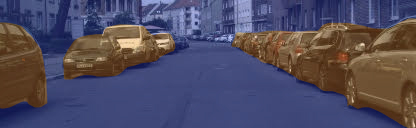}
\caption{Failure example 2, frame 1}
\end{subfigure}
\hspace*{\fill}
\begin{subfigure}{0.5\textwidth}
\includegraphics[width=\linewidth]{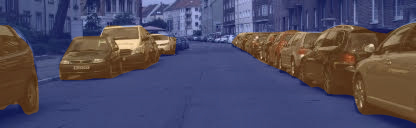}
\caption{Failure example 2, frame 2}
\end{subfigure}
\caption{Failure example 2: Segmentation network fails to segment out some cars at the end of the road on two consecutive frames.}
\label{fig:4_overlay}
\end{figure}

Consistency improves correctness, but does not guarantee it. A set of predictions can be consistent with one another, but not correct. A simple example is a misdetection of a static object. If the segmentation network fails to segment out the same object on two consecutive frames, the consistency loss will not penalize this failure. Fig.\ref{fig:1_overlay} and Fig.\ref{fig:4_overlay} shows some examples of failure cases that consistency was unable to fix. 

\section{Depth and Surface Normals}

To compute a consistency loss for the collective training of depth and normal prediction models we compute surface normals from the \emph{predicted depth map}, and penalize their deviation from the \emph{predicted normal map}, closely following the method in Ref.~\cite{depth2normals}. We first convert the the depth map to a 3D point cloud, using the inverse of the intrinsics matrix: 

\begin{equation}
\vec{r}_{ij} \equiv 
\begin{pmatrix}
x_{i,j} \\
y_{i,j} \\
z_{i,j}
\end{pmatrix}
= z_{i,j}
\cdot
\begin{pmatrix}
1/f_x & 0     & -x_0/f_x\\
0     & 1/f_y & -y_0/f_y \\
0 & 0 & 1 \textbf{}
\end{pmatrix}
\begin{pmatrix}
j \\
i \\
1
\end{pmatrix}
\end{equation}
where $f_x$ and $f_y$ denote the focal length, and $x_0$ and $y_0$ the principal point offset, $i$ and $j$ are the pixel coordinates along the height and the width of the image respectively, and $z_{ij}$ is the depth map evaluated at the pixel coordinates $(i, j)$. $\vec r_{ij}$ is a point in 3D space, in the camera coordinates, corresponding to pixel $(i,j)$.

We then compute the spatial derivatives of the depth map: 
\begin{equation}
\begin{split}
    (\partial x, \vec{r})_{i,j} = \vec{r}_{i, j+1} - \vec{r}_{i, j-1}\\
    (\partial y, \vec{r})_{i,j} = \vec{r}_{i+1, j} - \vec{r}_{i-1, j}
\end{split}
\end{equation}
To exclude depth discontinuities, we invalidate pixels where the spatial gradient of the depth relative to the depth itself is greater than a certain threshold $\beta$. 
To this end, we define a validity mask $v_{i, j}$: 
\begin{equation}
    V_{i,j} = (V_x)_{i,j} \cdot (V_y)_{i,j},
\end{equation}
where
\begin{equation}
\begin{split}
    (V_x)_{i,j} = 
    \begin{cases}
    1 & (\partial_{x}, z)_{i,j} < z_{i,j} \cdot \beta\\
    0 & \text{otherwise}
    \end{cases}\\    
    (V_y)_{i,j} = 
    \begin{cases}
    1 & (\partial_{y}, z)_{i,j} < z_{i,j} \cdot \beta\\
    0 & \text{otherwise}
    \end{cases}        
\end{split}    
\end{equation}

The validity mask $V_{i,j}$ is used to zero out the spatial gradients at depth discontinuities:
\begin{equation}
\begin{split}
    (\partial x, \vec{r'})_{i,j} = (\partial x, \vec{r})_{i,j} \cdot V_{i,j} \\
    (\partial y, \vec{r'})_{i,j} = (\partial y, \vec{r})_{i,j} \cdot V_{i,j} 
\end{split}
\end{equation}

We then compute the average spatial derivatives over a window of size $N\times N$ pixels around each pixel $(i, j)$: 
\begin{equation}
\begin{split}
    \overline{(\partial x, \vec{r})}_{i,j} = \frac{1}{N^2} \sum_{i',j}^{N}{(\partial x, \vec{r'})}_{i - \frac{N}{2} + i', j - \frac{N}{2} + j'} \\
    (\overline{\partial y, \vec{r})}_{i,j} = \frac{1}{N^2} \sum_{i',j}^{N}{(\partial y, \vec{r'})}_{i - \frac{N}{2} + i', j - \frac{N}{2} + j'}
\end{split}
\end{equation}
Note that we normalize by $N^2$, whereas the proper normalization would be by $\sum_{i', j'} V_{i - \frac{N}{2} + i', j - \frac{N}{2} + j'}$. However since the direction of the surface normal is insensitive to the norms of  $\overline{(\partial x, \vec{r})}_{i,j}$ and $\overline{(\partial y, \vec{r})}_{i,j}$, as opposed to their directions, this is immaterial.

To obtain the surface normal, we calculate the cross product of $\overline{(\partial x, \vec{r})}_{i,j}$ and $\overline{(\partial y, \vec{r})}_{i,j}$,
\begin{equation}
    (\vec{n_d})_{i,j} =  \overline{(\partial x, \vec{r})}_{i,j} \times (\overline{\partial y, \vec{r})}_{i,j} 
\end{equation}
and then normalize it, to obtain the normalized surface normal:
\begin{equation}
    (\hat{n_d})_{i,j} = (\vec{n_d})_{i,j} / \left\lVert (\vec{n_d})_{i,j} \right\rVert
\end{equation}
The consistency is then computed as 

\begin{equation}
    \mathcal{L}_{consistency} = \textrm{cosine\_distance}(\hat{n_d}, \hat{n}_p),
\end{equation}
where $\hat{n_d}$ is the computed surface normals from the inferred depth as described in this section, and $\hat{n_p}$ is the normal map predicted from the normal prediction network.

\begin{figure} [t]
\centering
\includegraphics[width=\linewidth]{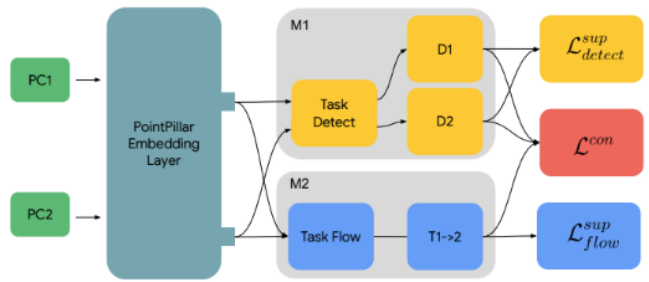}
\caption{3D Detection and flow estimation joint training: we use parts of Waymo Open Dataset~\cite{sun2019scalability} to supervise the training of a single-frame 3D detector (M1) and multi-frame flow estimator (M2) and apply a motion consistency loss on a set of unlabeled data. }
\label{fig:domain_adapt_point_clounds}
\end{figure}

\section{3D Object Detection in Point Clouds in Time}

To add more detail, we visualize the joint training of 3D Object detection and flow in
Figure~\ref{fig:domain_adapt_point_clounds}. 
Two consecutive Point Clouds are considered PC1 and PC2, which are processed by a Point Pillar embedding layer~\cite{lang2019pointpillars}. The first module (M1) provides the 3D detections D1 and D2 for the two point clouds, respectively, whereas the second module (M2) computes the flow $T1 \rightarrow 2$, which is limited to boxes only. Each of these modules applies its respective supervision, whereas the consistency loss imposes that the detected boxes from the static point cloud should match the detections from the previous static point cloud as transformed by the predicted flow (Figure~\ref{fig:domain_adapt_point_clounds}).

\end{document}